\pdfoutput=1

\documentclass[11pt]{article}

\usepackage{acl}
\usepackage{times}
\usepackage{latexsym}

\usepackage[T1]{fontenc}

\usepackage[utf8]{inputenc}

\usepackage{microtype}
\usepackage{times}
\usepackage{latexsym}
\usepackage{multirow}
\usepackage{url}

\usepackage{adjustbox}
\usepackage{graphicx}
\usepackage[scaled=.90]{helvet}
\usepackage{amsfonts}
\usepackage{float}
\restylefloat{table}
\usepackage{booktabs}
\usepackage{amsmath}
\usepackage{enumitem}
\usepackage{multirow}
\usepackage{subcaption}
\usepackage{cleveref}
\crefformat{section}{\S#2#1#3}
\crefformat{subsection}{\S#2#1#3}

\usepackage[normalem]{ulem}

\usepackage{caption}

\title{Improving Robustness of Retrieval Augmented Translation via \\ Shuffling of Suggestions}

\author{Cuong Hoang\thanks{\hspace{1.5mm}Work done while the authors were at AWS AI Labs.}, Devendra Sachan\footnotemark[1], Prashant Mathur, 
Brian Thompson, Marcello Federico\\ 
AWS AI Labs \\
\href{mailto:pramathu@amazon.com}{pramathu@amazon.com}}

\date{}

\begin{document}
\maketitle
\begin{abstract}
Several recent studies have reported dramatic performance improvements in neural machine translation (NMT) by augmenting translation at inference time with fuzzy-matches retrieved from a translation memory (TM).
However, these studies all operate under the assumption that the TMs available at test time are highly relevant to the testset. 
We demonstrate that for existing retrieval augmented translation methods, using a TM with a domain mismatch to the test set can result in substantially worse performance  compared to not using a TM at all. 
We propose a simple method to expose fuzzy-match NMT systems during training and show that it results in a system that is much more tolerant (regaining up to $5.8$ BLEU) to inference with TMs with domain mismatch. {Also, the model is still competitive to the baseline when fed with suggestions from relevant TMs.} %
\end{abstract}

\section{Introduction}
\label{sec:intro}
Retrieval Augmented Translation (RAT) refers to a paradigm 
which combines a translation model~\cite{NIPS2017_7181} with an external retriever module~\cite{RATsurvey}.
The \textit{retrieval module} (e.g. BM25 ranker \cite{robertson-bm25-2009} or a neural retriever \cite{cai-etal-2021-neural,sachan-etal-2021-end}) takes each source sentence as input and retrieves the top-$k$ most similar target translations from a Translation Memory (TM)~\cite{farajian-etal-2017-multi,DBLP:journals/corr/GuWCL17,bulte-tezcan-2019-neural}. 
The \textit{translation module} then encodes the input along with the top-$k$ fuzzy-matches, either by appending the suggestions to the input \cite{bulte-tezcan-2019-neural,xu-etal-2020-boosting} or using separate encoders for input and suggestions \cite{he-etal-2021-fast,cai-etal-2021-neural}. 
Decoder then either learns to copy or carry over the stylistics from the suggestions while generating the translation. 
In this work, we focus only on the translation module of this paradigm.

In existing literature, 
inference with RAT models have typically assumed that TMs are \textit{domain-matched} i.e. the test set is of the same domain as the translation memory. 
Many works (e.g. ~\citet{bulte-tezcan-2019-neural}, \citet{xu-etal-2020-boosting} and \citet{cai-etal-2021-neural}) have reported dramatic performance improvements under such setting. However, it is not clear how the models perform when there is a domain-mismatch between the TM and the test set.

In this work, we focus on the problem where the assumption of TM being domain-matched with test set does not hold. We explore the conditions where models are provided suggestions from a TM, but not from the same domain as the test set.
We show that RAT models suffer performance drop if fed with suggestions coming from less relevant TMs.

This finding is especially important from a usability standpoint. 
Often a translator may pick a TM as a best fit for a translation job if an ideal TM is not available or created just yet e.g. \texttt{IT} domain TM is picked for \texttt{Patent} translation job because the domains are closer and there no available TM for \texttt{Patent}. 
This could lead to multiple issues like ambiguous terminologies or multiple meanings across same context \cite{jalili-sabet-etal-2016-tmop,10.1007/s10590-016-9183-x}).
RAT model leveraging such (mismatched) TM ends up producing worse quality translation than a standard MT system.
Therefore, it is desirable that RAT models not only improve translation with suggestions from relevant TMs, but also be more robust to suggestions from less relevant TMs.

To this end, we propose an enhancement to the training of RAT models with a simple shuffling method to mitigate this problem. Instead of
always using $k$ most-relevant fuzzy-matches in training,
our method \textit{randomly samples} $k$ from a larger list 
(e.g. randomly sample 3 sentences from the top-$10$ matches). 
Our hypothesis is that if we systematically provide only top similar suggestions during training, the model will overly rely on the suggestions and simply copy the tokens in them. By shuffling the retrieved results, we ensure suggestions to be less similar to the input and train the system to be more robust to less relevant suggestions at test time. Our experiment results show that the model trained with shuffling of suggestions outperforms the standard RAT model by up to $+5.8$ BLEU score on average when suggestions come from less relevant TMs while dropping $-0.15$ BLEU on average when suggestions come from relevant TMs. 

To the best of our knowledge, this is the first work to consider the robustness of RAT methods, which we believe is critical for acceptance by human translators. 

\section{Related Work}
\label{sec:additional}

RAT is a form of domain adaptation, which is often achieved via continued training in NMT~\cite{freitag2016fast,luong-manning-2015-stanford}. 
However, RAT differs from standard domain adaptation techniques like continued training in that it is \textit{online}, that is, the model is not adapted during training and instead domain adaptation occurs at inference time. This makes RAT better suited for some real-world applications e.g. a single server with a single model loaded in memory can serve hundreds or thousands of users with custom translations, adapted to their unique TM. 
Other works have considered online adaptation outside the context of RAT, including \citet{vilar-2018-learning}, who propose Learning Hidden Unit Contributions  \cite{swietojanski2016learning} as a compact way to store many adaptations of the same general-domain model.

Previous works in retrieval augmented translation have mainly explored 
aspects of filtering fuzzy-matches by applying similarity thresholds  \cite{Xia_Huang_Liu_Shi_2019,xu-etal-2020-boosting}, leveraging word alignment information \cite{zhang-etal-2018-guiding,xu-etal-2020-boosting,he-etal-2021-fast} or re-ranking with additional score (e.g. word overlapping) \cite{DBLP:conf/aaai/GuWCL18,zhang-etal-2018-guiding}. 
Our approach do not make use of any filtering and as such do not require any ad-hoc optimization.
Our work is related to the use of $k$-nearest-neighbor for NMT  \cite{khandelwal2021nearest,zheng-etal-2021-adaptive} but it is less expensive and does not require storage and search over a large data store of context representations and corresponding target tokens \cite{DBLP:journals/corr/abs-2105-14528}.

Our work also relates to work in \textit{offline} adaptation which has addressed catastrophic forgetting 
of general domain knowledge during domain adaptation \cite{thompson-etal-2019-overcoming} via 
ensembling in-domain and out-of-domain models \cite{freitag2016fast}, 
mixing of in-domain and out-of-domain data during adaptation \cite{chu-etal-2017-empirical},
multi-objective learning and multi-output learning \cite{dakwale2017finetuning},
and 
elastic weight consolidation \cite{kirkpatricEWC, thompson-etal-2019-overcoming, saunders-etal-2019-domain}, 
or combinations of these techniques \cite{hasler-etal-2021-improving}.

\section{RAT with Shuffling}
\subsection{Retrieval Module}
We use Okapi BM25~\cite{robertson-bm25-2009}, a classical retrieval algorithm that performs search by computing lexical matches of the query with all sentences in the evidence, to obtain top-ranked sentences for each input.\footnote{To enable fast retrieval, we leverage the implementation provided by the ElasticSearch library, which can be found at \url{https://github.com/elastic/elasticsearch-py}.} Specifically, we built an index using source sentences of a TM. For every input, we collect top-$k$ (i.e. $k=\{1, 2, 3, 4, 5\}$  in our experiments) similar source side sentences and then use their corresponding target side sentences as fuzzy-matches.%

\subsection{Shuffling suggestions}\label{model_section}
We propose to relax the use of top-$k$ relevant suggestions during training by training RAT model with $k$ fuzzy-matches randomly sampled from a larger list. In our experiments, we sample $k$ from the top-$10$ matches, where top-$10$ is chosen based on our preliminary experiments. 

By shuffling retrieval fuzzy-matches, we ensure suggestions to be less similar to the target reference. With that, we expect the model learns to be more selective in using the suggestions for translation and thus be more robust to less relevant suggestions at test time. In fact, training models with noisy data has been shown to improve model's robustness to irrelevant data~\cite{DBLP:conf/iclr/BelinkovB18}.

\section{Data, Models \& Experiments}
\subsection{Data}
We conduct experiments in two language directions: En-De with five domain-specialized TMs\footnote{\texttt{Medical}, \texttt{Law}, \texttt{IT}, \texttt{Religion} and \texttt{Subtitles}.} and En-Fr with seven domain-specialized TMs.\footnote{\texttt{News}, \texttt{Medical}, \texttt{Bank}, \texttt{Law}, \texttt{IT}, \texttt{TED} and \texttt{Religion}.} En-De data is taken from \newcite{aharoni2020unsupervised}, which is a re-split version of the multi-domain data set from \newcite{koehn-knowles-2017-six} while En-Fr data set is taken from the multi-domain data set of \newcite{10.1162/tacl_a_00351}. We concatenate all parallel training data from multiple domains for training and similarly for development sets.

\subsection{Models and Training}
We experiment with a common RAT model used in \newcite{bulte-tezcan-2019-neural} and \newcite{xu-etal-2020-boosting}. The model encodes the input along with the top-$k$ fuzzy-matches by appending the suggestions to the input. We follow the recipe of \newcite{NIPS2017_7181} during training. We report translation quality with BLEU scores computed via \texttt{Sacrebleu}~\cite{post-2018-call}.\footnote{Details of all hyper-parameters of training and evaluation are presented in Appendix~\ref{appendix:params}.} We use \texttt{compare-mt} \cite{neubig-etal-2019-compare} for
significance testing with \texttt{bootstrap = 1000} and
\texttt{prob\_thresh = 0.05}. 

\subsection{Inference with TMs}
We run inference with RAT models using top-k suggestions coming from \textit{relevant} and \textit{less relevant} TMs. Specifically, for every input sentence in a test set from one specific domain, less relevant suggestions are retrieved from all TMs but not the TM coming from the same domain to the test set. Taking En-De and \texttt{IT} domain test set as an example, less relevant suggestions are collected from all TMs coming from \texttt{Medical}, \texttt{Law}, \texttt{Religion} and \texttt{Subtitles} but not \texttt{IT} domain.
\begin{table}[ht!]
\small
\centering
\begin{tabular}{l|l|c} 
\multicolumn{3}{c}{En $\rightarrow$ De}\\ \hline
Model &TM&\textbf{AVER}\\ \hline
Transformer &-&37.61\\\hline
RAT &\multirow{2}{*}{Less relevant}&32.89\\
RAT + Shuf.&&\textbf{35.28/+2.39}\\\hline
RAT &\multirow{2}{*}{Relevant}&41.34\\
RAT + Shuf.&&\textbf{41.19/-0.15}\\ \hline
\end{tabular}
\begin{tabular}{l|l|c}
\multicolumn{3}{c}{En $\rightarrow$ Fr}\\ \hline
Model &TM&\textbf{AVER}\\ \hline
Transformer&-&55.86\\ \hline
RAT &\multirow{2}{*}{Less relevant}&44.39\\
RAT + Shuf.&&\textbf{49.20/+5.81} \\ \hline
RAT &\multirow{2}{*}{Relevant}&56.59\\
RAT + Shuf.&&\textbf{56.44/-0.15}\\ \hline
\end{tabular}
\caption{Average BLEU scores of models across top-k values and across domains. The model trained with shuffling is more tolerant to domain mismatch in TMs while still being competitive when fed with suggestions from relevant TMs.}
\label{mainresults_average}
\end{table}
\begin{table*}[ht]
\small
\centering
\begin{tabular}{l|lllllll} 
\multicolumn{8}{c}{En $\rightarrow$ De}\\ \hline
Model & Top-k & IT & LAW & REL & MED & SUBT & \textbf{AVER}\\  \hline
Baseline  &-& 39.22&53.06&23.03&48.48&24.24&37.61\\  \hline
RAT &\multirow{2}{*}{k=2}&\textbf{41.16}&44.41&16.48&41.95&23.84&33.57\\
RAT + Shuf.&&40.81&\textbf{49.37*}&\textbf{21.01*}&\textbf{46.85*}&23.92&\textbf{36.39}\\\hline
RAT &\multirow{2}{*}{k=3}&\textbf{41.03}&43.61&15.11&40.35&23.35&32.69\\
RAT + Shuf.&&40.33&\textbf{46.52*}&\textbf{18.37*}&\textbf{44.66*}&\textbf{24.18}&\textbf{34.81}\\\hline
RAT &\multirow{2}{*}{k=4}&40.91&42.98&14.39&39.09&23.56&32.19\\
RAT + Shuf.&&\textbf{42.0}&\textbf{46.4*}&\textbf{18.23*}&\textbf{43.65*}&\textbf{24.02}&\textbf{34.86}\\\hline
\end{tabular}
\begin{tabular}{l|lllllllllll}
\multicolumn{10}{c}{En $\rightarrow$ Fr}\\ \hline
Model & Top-k& LAW & MED &IT & NEWS & BANK & REL & TED & \textbf{AVER}\\ \hline
Baseline &-& 65.58&57.92&48.89&34.43&57.32&86.87&40.03&55.86\\\hline
RAT & \multirow{2}{*}{k=2}&56.89&51.36&42.18&33.15&47.42&49.62&39.09&45.67\\
RAT + Shuf.&&\textbf{60.20*}&\textbf{52.66*}&\textbf{45.55*}&\textbf{34.12*}&\textbf{50.36*}&\textbf{81.76*}&\textbf{39.85}&\textbf{52.07}
\\\hline
RAT & \multirow{2}{*}{k=3}&56.48&50.31&40.41&33.25&45.32&43.61&38.31&43.96\\
RAT + Shuf.&& \textbf{57.81*}&\textbf{51.89*}&\textbf{42.78*}&\textbf{33.51}&\textbf{48.08*}&\textbf{73.11*}&\textbf{38.75}&\textbf{49.42}\\\hline
RAT & \multirow{2}{*}{k=4}&54.93&49.98&39.77&\textbf{33.05}&45.03&39.26&37.75&42.82\\
RAT + Shuf. && \textbf{56.33*}&\textbf{50.92*}&\textbf{42.07*}&32.94&\textbf{46.58}&\textbf{60.31*}&\textbf{38.48}&\textbf{46.80}\\
\hline
\end{tabular}
\caption{Model comparison with \textbf{less relevant} suggestions. The best BLEU for models with a specific top-k value is highlighted in bold, and "*" indicates whether the models are statistical significance difference with the others.}\label{mainresults_less relevant}
\end{table*}
\begin{table*}[ht]
\small
\centering
\begin{tabular}{l|lllllll} 
\multicolumn{8}{c}{En $\rightarrow$ De}\\ \hline
Model & Top-k & IT & LAW & REL & MED & SUBT & \textbf{AVER}\\  \hline
Baseline  &-& 39.22&53.06&23.03&48.48&24.24&37.61\\ \hline
RAT & \multirow{2}{*}{k=2}&\textbf{41.52}&\textbf{57.86}&30.67&\textbf{53.52}&23.83&\textbf{41.48}\\
RAT + Shuf.&&40.78&57.82&\textbf{30.87}&53.34&\textbf{24.07}&41.38\\\hline
RAT &\multirow{2}{*}{k=3}&\textbf{41.43}&\textbf{57.67}&\textbf{31.24}&\textbf{53.05*}&23.66&\textbf{41.41}\\
RAT + Shuf.&&40.90&56.98&30.65&51.36&\textbf{24.24}&40.83\\\hline
RAT &\multirow{2}{*}{k=4}&41.35&57.54&31.7&52.43&23.53&41.31\\
RAT + Shuf.&&\textbf{42.24}&\textbf{58.90*}&\textbf{33.02*}&\textbf{53.7}&\textbf{23.94}&\textbf{42.36}\\\hline
\end{tabular}
\begin{tabular}{l|lllllllllll}
\multicolumn{10}{c}{En $\rightarrow$ Fr}\\ \hline
Model & Top-k& LAW & MED &IT & NEWS & BANK & REL & TED & \textbf{AVER}\\ \hline
Baseline &-& 65.58&57.92&48.89&34.43&57.32&86.87&40.03&55.86\\\hline
RAT& \multirow{2}{*}{k=2}&\textbf{70.25*}&\textbf{60.20*}&\textbf{50.28}&33.95&\textbf{58.65*}&83.99&40.32&\textbf{56.81}\\
RAT + Shuf.&&69.45&59.42&49.41&\textbf{34.38}&57.46&\textbf{84.15}&\textbf{40.63}&56.41
\\\hline
RAT&\multirow{2}{*}{k=3}&\textbf{70.70}&\textbf{61.02}&\textbf{52.57}&33.61&\textbf{60.67}&84.84&39.47&57.55\\
RAT + Shuf.&&{70.03}&{60.41}&51.69&\textbf{33.64}&60.41&\textbf{85.15}&\textbf{39.66}& \textbf{57.28}
\\\hline
RAT &\multirow{2}{*}{k=4}&\textbf{70.07}&\textbf{60.23*}&\textbf{50.28}&33.20&\textbf{57.5}&84.65&39.37&\textbf{56.47}\\
RAT + Shuf.&& 69.73&59.03&49.4&\textbf{33.32}&56.63&\textbf{85.29}&\textbf{39.78}&56.17\\\hline
\end{tabular}
\caption{Model comparison with \textbf{relevant} suggestions. The
best BLEU for models with a specific top-k value is highlighted in bold, and "*" indicates whether the models are statistical significance difference with the others.}\label{mainresults_relevant}
\end{table*}
\section{Results}\label{multidomain}
\subsection{Main results}
We experiment with comprehensive top-k values of $\{1, 2, 3, 4, 5\}$ and compare models when fed with suggestions from both relevant and less relevant TMs. The summary results across top-k values and across domains are collected in Table~\ref{mainresults_average} for both En-De and En-Fr.
When the standard RAT models are fed with less relevant suggestions, we see a drop of 12.5\% (37.61~$\rightarrow$~32.89) for En-De and 20.5\% (55.86~$\rightarrow$~44.39) for En-Fr against standard Transformer based MT system. This shows how brittle the RAT models are when the assumption -  TMs are of same domain as test set - breaks.  
Our approach with shuffling closes the gap by 50\% from 4.72 (37.61~$\rightarrow$~32.89) to 2.39 (37.61~$\rightarrow$~35.28) for En-De and 54.7\% similarly for En-Fr when RAT models are fed with less relevant suggestions while dropping 0.15 when the suggestions are from relevant TMs. This is an acceptable trade-off from the point of view of a user because of the significant gains observed in the less relevant TM setting.

Details with a subset of top-$k$ values of $\{2, 3, 4\}$ are in Table~\ref{mainresults_less relevant} (with less relevant TMs) and Table~\ref{mainresults_relevant} (with relevant TMs).\footnote{We report a subset of results due to space constraint. Comprehensive results are shown in Appendix~\ref{appendix:results}.}
The improvements with shuffling are consistent for all $k$s when models are fed less relevant TMs and clearly support that shuffling suggestions during training significantly improves robustness of RAT models.
In Table~\ref{mainresults_relevant}, where the models are fed with relevant TMs, we observe a drop in translation quality in some cases (e.g. \texttt{MED} domain in En-De with k=3) but on an average performance of the models across domains remains competitive. %

\begin{table}[ht!]
\small
\begin{tabular}{l|llllllllllll} 
Model &IT&LAW&REL&MED&SUBT\\  \cline{0-8}
\multicolumn{6}{c}{Less Relevant TM}\\ \cline{0-5}
RAT &24.82&17.81&19.25&14.73&20.69\\
+ Shuf.&23.51*&17.04*&17.74*&13.42*&20.05*\\\hline
\multicolumn{6}{c}{Relevant TM}\\ \cline{0-5}
RAT &27.01&40.75&36.37&39.45&20.67&\\
+ Shuf.&25.63*&39.05*&34.33*&37.46*&19.99*&\\\hline
\end{tabular}
\caption{Average percentage of suggestion tokens that appear in MT outputs produced by RAT models across five En-De test sets, with both relevant and less relevant TMs. ``*'' indicates statistical significance difference.}\label{tab:percentage}
\end{table}

\subsection{Usage of Suggestions}
To understand the direct impact of suggestions on translation, we looked at the percentage of tokens in suggestions that appear in output produced by model. Average percentage of overlapping tokens for each test set is in Table~\ref{tab:percentage}, with $k$ = 3 and En-De as a case study. 

Our results show a consistently significantly less use of less relevant TMs by RAT models trained with shuffling method. This clearly support our postulation that RAT models trained with shuffling are more selective in using suggestions for translation, which explains the robustness of the model when it is run inference with less relevant suggestions.

As the model is more selective, it also tends to less use of suggestions from relevant TMs as in Table~\ref{tab:percentage}. This explains the translation accuracy drop in some cases (e.g. \texttt{MED} domain in En-De with k=3) when fed with relevant suggestions in exchange of the robustness.

\section{Conclusion}
We show that standard RAT systems are not robust to domain-mismatch at test time. We proposed a training procedure of shuffling fuzzy-match suggestions to mitigate the problem. Our experiment results show that the model trained with shuffling significantly outperforms the one without shuffling, across multiple domains on two language pairs when suggestions come from less relevant TMs at inference time. We also show that RAT with shuffling is still competitive to models without shuffling across all domains when fed with relevant suggestions. Further analysis shows that the model trained with shuffling is more selective in using suggestions for translation, which explains the increased robustness.

\bibliography{main.bbl}
\bibliographystyle{acl_natbib}

\clearpage

\addcontentsline{toc}{section}{Appendices}
\renewcommand{\thesubsection}{\Alph{subsection}}
\section*{Appendix}

\subsection{Model Parameters}\label{appendix:params}
We employed Transformer big architecture with 6 encoder and 6 decoder layers. Hidden size was set to  $1024$ and maximum input length to $1024$ tokens. We employed a joint source-target language subword vocabulary of size $32K$ using Sentencepiece algorithm~\cite{DBLP:conf/emnlp/KudoR18}.

We use the Adam optimizer~\cite{KingmaB14} with $\beta_1 = 0.9$, $\beta_2 = 0.98$ and $\epsilon = 10^{-9}$; and (ii) increase the learning rate linearly for the first $4K$ training steps and decrease it thereafter; (iii) use batch size of $32K$ source tokens and $32K$ target tokens. Checkpoints are saved after every $10K$ iterations during training. We use dropout of $0.1$ and label-smoothing of $0.1$. We train models with maximum of $300K$ iterations.

For evaluation we use Sacrebleu with hash value of \texttt{nrefs:1|case:mixed|eff:no|tok:13a|\\smooth:exp|version:2.0.0}.

\subsection{Full results with $k=\{1, 2, 3, 4, 5\}$}\label{appendix:results}
Details with full set of top-$k$ values of $\{1, 2, 3, 4, 5\}$ are in Table~\ref{amainresults_less relevant} (i.e. with less relevant TMs) and Table~\ref{amainresults_relevant} (i.e. with relevant TMs).

\begin{table*}[ht]
\small
\begin{tabular}{l|llllllllllll} 
\multicolumn{9}{c}{En $\rightarrow$ De}\\ \cline{0-8}
Model &TM&Top-k&IT&LAW&REL&MED&SUBT&\textbf{AVER}\\  \cline{0-8}
Baseline  &-&-& 39.22&53.06&23.03&48.48&24.24&37.61\\  \cline{0-8}
RAT &\multirow{10}{*}{Less relevant}&\multirow{2}{*}{k=1}&\textbf{40.82}&45.61&17.98&42.65&24.08&34.22\\
RAT + Shuf.&&&39.56&\textbf{50.89*}&\textbf{21.6*}&\textbf{47.8*}&\textbf{24.33}&\textbf{36.84}\\\cline{3-9}
RAT &&\multirow{2}{*}{k=2}&\textbf{41.16}&44.41&16.48&41.95&23.84&33.57\\
RAT + Shuf.&&&40.81&\textbf{49.37*}&\textbf{21.01*}&\textbf{46.85*}&23.92&\textbf{36.39}\\\cline{3-9}
RAT &&\multirow{2}{*}{k=3}&\textbf{41.03}&43.61&15.11&40.35&23.35&32.69\\
RAT + Shuf.&&&40.33&\textbf{46.52*}&\textbf{18.37*}&\textbf{44.66*}&\textbf{24.18}&\textbf{34.81}\\\cline{3-9}
RAT &&\multirow{2}{*}{k=4}&40.91&42.98&14.39&39.09&23.56&32.19\\
RAT + Shuf.&&&\textbf{42.0}&\textbf{46.4*}&\textbf{18.23*}&\textbf{43.65*}&\textbf{24.02}&\textbf{34.86}\\\cline{3-9}
RAT &&\multirow{2}{*}{k=5}&40.83&41.93&13.8&38.55&23.67&31.76\\
RAT + Shuf.&&&\textbf{41.04}&\textbf{44.41*}&\textbf{16.25*}&\textbf{41.57*}&\textbf{24.12}&\textbf{33.48}\\\hline
\multicolumn{9}{c}{En $\rightarrow$ Fr}\\ \hline
Model &TM&Top-k& LAW & MED &IT & NEWS & BANK & REL & TED & \textbf{AVER}\\ \hline
Baseline &-&-& 65.58&57.92&48.89&34.43&57.32&86.87&40.03&55.86\\\hline
RAT&\multirow{10}{*}{Less relevant}&\multirow{2}{*}{k=1}&58.3&51.54&44.99&33.51&48.49&59.62&39.07&47.93\\
RAT + Shuf.&&&\textbf{60.97*}&\textbf{53.48*}&\textbf{45.95}&\textbf{33.76}&\textbf{51.2}&\textbf{82.37*}&\textbf{39.73}&\textbf{52.49}\\\cline{3-11}
RAT&&\multirow{2}{*}{k=2}&56.89&51.36&42.18&33.15&47.42&49.62&39.09&45.67\\
RAT + Shuf.&&&\textbf{60.20*}&\textbf{52.66*}&\textbf{45.55*}&\textbf{34.12*}&\textbf{50.36*}&\textbf{81.76*}&\textbf{39.85}&\textbf{52.07}
\\\cline{3-11}
RAT&&\multirow{2}{*}{k=3}&56.48&50.31&40.41&33.25&45.32&43.61&38.31&43.96\\
RAT + Shuf.&&&\textbf{57.81*}&\textbf{51.89*}&\textbf{42.78*}&\textbf{33.51}&\textbf{48.08*}&\textbf{73.11*}&\textbf{38.75}&\textbf{49.42}\\\cline{3-11}
RAT&&\multirow{2}{*}{k=4}&54.93&49.98&39.77&\textbf{33.05}&45.03&39.26&37.75&42.82\\
RAT + Shuf.&&&\textbf{56.33*}&\textbf{50.92*}&\textbf{42.07*}&32.94&\textbf{46.58}&\textbf{60.31*}&\textbf{38.48}&\textbf{46.80}
\\\cline{3-11}
RAT&&\multirow{2}{*}{k=5}&54.36&49.91&39&32.2&44.38&33.88&37.23&41.57\\
RAT + Shuf.&&&\textbf{55.43*}&\textbf{50.41}&\textbf{40.51*}&\textbf{32.81}&\textbf{45.26}&\textbf{54.02*}&\textbf{37.91}&\textbf{45.19}\\\cline{3-11}
\end{tabular}
\caption{Full results of comparing models with \textit{less relevant} fuzzy-matches. Here, top-k values are with $\{1, 2, 3, 4, 5\}$. The
best BLEU for models with a specific top-k value is highlighted in bold, and "*" indicates whether the models are statistical significance difference with the others.}\label{amainresults_less relevant}
\end{table*}
\begin{table*}[ht]
\small
\begin{tabular}{l|llllllllllll} 
\multicolumn{9}{c}{En $\rightarrow$ De}\\ \cline{0-8}
Model &TM&Top-k&IT&LAW&REL&MED&SUBT&\textbf{AVER}\\  \cline{0-8}
Baseline  &-&-& 39.22&53.06&23.03&48.48&24.24&37.61\\  \cline{0-8}
RAT &\multirow{10}{*}{Relevant}&\multirow{2}{*}{k=1}&\textbf{41.24*}&\textbf{57.66*}&\textbf{29.11*}&\textbf{53.73*}&\textbf{24.04}&\textbf{41.16}\\
RAT + Shuf.&&&39.55&55.84&25.92&50.8&24.30&39.28\\\cline{3-9}
RAT &&\multirow{2}{*}{k=2}&\textbf{41.52}&\textbf{57.86}&30.67&\textbf{53.52}&23.83&\textbf{41.48}\\
RAT + Shuf.&&&40.78&57.82&\textbf{30.87}&53.34&\textbf{24.07}&41.38\\\cline{3-9}
RAT &&\multirow{2}{*}{k=3}&\textbf{41.43}&\textbf{57.67}&\textbf{31.24}&\textbf{53.05*}&23.66&\textbf{41.41}\\
RAT + Shuf.&&&40.90&56.98&30.65&51.36&\textbf{24.24}&40.83\\\cline{3-9}
RAT &&\multirow{2}{*}{k=4}&41.35&57.54&31.7&52.43&23.53&41.31\\
RAT + Shuf.&&&\textbf{42.24}&\textbf{58.90*}&\textbf{33.02*}&\textbf{53.7}&\textbf{23.94}&\textbf{42.36}\\\cline{3-9}
RAT &&\multirow{2}{*}{k=5}&40.58&57.2&32.1&53.21&23.69&41.36\\
RAT + Shuf.&&&\textbf{41.43}&\textbf{58.32*}&\textbf{33.07}&\textbf{53.89}&\textbf{23.78}&\textbf{42.10}\\\hline
\multicolumn{9}{c}{En $\rightarrow$ Fr}\\ \hline
Model &TM&Top-k& LAW & MED &IT & NEWS & BANK & REL & TED & \textbf{AVER}\\ \hline
Baseline &-&-& 65.58&57.92&48.89&34.43&57.32&86.87&40.03&55.86\\\hline
RAT&\multirow{10}{*}{Relevant}&\multirow{2}{*}{k=1}&\textbf{70.39*}&\textbf{59.45}&\textbf{51.77*}&34.07&\textbf{57.42}&81.15&39.93&56.31\\
RAT + Shuf.&&&69.2&58.83&49.75&\textbf{34.26}&57.29&\textbf{84.92*}&\textbf{40.16}&\textbf{56.34}
\\\cline{3-11}
RAT&&\multirow{2}{*}{k=2}&\textbf{70.25*}&\textbf{60.20*}&\textbf{50.28}&33.95&\textbf{58.65*}&83.99&40.32&\textbf{56.81}\\
RAT + Shuf.&&&69.45&59.42&49.41&\textbf{34.38}&57.46&\textbf{84.15}&\textbf{40.63}&56.41
\\\cline{3-11}
RAT&&\multirow{2}{*}{k=3}&\textbf{70.70}&\textbf{61.02}&\textbf{52.57}&33.61&\textbf{60.67}&84.84&39.47&57.55\\
RAT + Shuf.&&&{70.03}&{60.41}&51.69&\textbf{33.64}&60.41&\textbf{85.15}&\textbf{39.66}& \textbf{57.28}
\\\cline{3-11}
RAT&&\multirow{2}{*}{k=4}&\textbf{70.07}&\textbf{60.23*}&\textbf{50.28}&33.20&\textbf{57.5}&84.65&39.37&\textbf{56.47}\\
RAT + Shuf.&&&69.73&59.03&49.4&\textbf{33.32}&56.63&\textbf{85.29}&\textbf{39.78}&56.17\\\cline{3-11}
RAT&&\multirow{2}{*}{k=5}&68.99&59.39&\textbf{49.61}&32.85&\textbf{56.92}&83.88&39.14&55.83\\
RAT + Shuf.&&&\textbf{69.55}&\textbf{59.51}&49.60&\textbf{33.37}&56.45&\textbf{84.13}&\textbf{39.44}&\textbf{56.01}
\\\cline{3-11}
\end{tabular}
\caption{Full results of comparing models with \textit{relevant} fuzzy-matches. Here, top-k values are with $\{1, 2, 3, 4, 5\}$. The
best BLEU for models with a specific top-k value is highlighted in bold, and "*" indicates whether the models are statistical significance difference with the others.}\label{amainresults_relevant}
\end{table*}

\end{document}